\documentclass{article}
\usepackage{arxiv}
\usepackage{graphicx}
\usepackage{hyperref}
\usepackage{float}
\usepackage{subfig}
\usepackage{caption}
\usepackage{tabularx} 
\usepackage{array}
\usepackage{multirow}
\usepackage{booktabs}
\usepackage{threeparttable}
\usepackage{tablefootnote}
\usepackage[table,xcdraw]{xcolor}
\usepackage{url}
\usepackage{comment}
\usepackage[utf8]{inputenc} 
\usepackage[T1]{fontenc}    
\usepackage{hyperref}      
\usepackage{url}            
\usepackage{booktabs}       
\usepackage{amsfonts}       
\usepackage{nicefrac}       
\usepackage{microtype}      
\usepackage{lipsum}		
\usepackage{graphicx}
\usepackage{doi}
\usepackage{float}
\usepackage{subfig}
\usepackage{caption}
\usepackage{tabularx} 
\usepackage{array}
\usepackage{multirow}
\usepackage{booktabs}
\usepackage{threeparttable}
\usepackage{tablefootnote}
\usepackage[table,xcdraw]{xcolor}
\usepackage{url}
\usepackage{comment}
\usepackage{changepage}
\usepackage[backend=biber, style=numeric, sorting=nyt]{biblatex}
\title{Improving Address Matching\\ using Siamese Transformer Networks}

\author{ \href{https://orcid.org/0000-0001-5987-0789}{\includegraphics[scale=0.06]{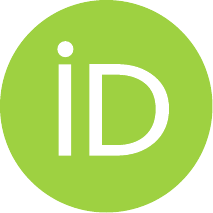}\hspace{1mm}André V. Duarte}\\
	\And
	\href{https://orcid.org/0000-0001-8638-5594}{\includegraphics[scale=0.06]{orcid.pdf}\hspace{1mm}Arlindo L. Oliveira} \\
}

\renewcommand{\headeright}{ preprint }
\renewcommand{\undertitle}{ \; }
\renewcommand{\shorttitle}{Improving Address Matching using Siamese Transformer Networks}

\renewenvironment{abstract}
{
  \begin{center}
    \large \bfseries \scshape Abstract
  \end{center}
  \begin{adjustwidth}{6em}{6em} 
}
{
  \end{adjustwidth}
}

\begin{document}
\maketitle              

\begin{center}
    Instituto Superior Técnico / INESC-ID \\
	\texttt{\{andre.v.duarte, arlindo.oliveira\}@tecnico.ulisboa.pt}
\end{center}

\vskip 0.3in

\begin{abstract}

Matching addresses is a critical task for companies and post offices involved in the processing and delivery of packages. The ramifications of incorrectly delivering a package to the wrong recipient are numerous, ranging from harm to the company's reputation to economic and environmental costs. This research introduces a deep learning-based model designed to increase the efficiency of address matching for Portuguese addresses. The model comprises two parts: (i) a bi-encoder, which is fine-tuned to create meaningful embeddings of Portuguese postal addresses, utilized to retrieve the top 10 likely matches of the un-normalized target address from a normalized database, and (ii) a cross-encoder, which is fine-tuned to accurately rerank the 10 addresses obtained by the bi-encoder. The model has been tested on a real-case scenario of Portuguese addresses and exhibits a high degree of accuracy, exceeding 95\% at the door level. When utilized with GPU computations, the inference speed is about 4.5 times quicker than other traditional approaches such as BM25. An implementation of this system in a real-world scenario would substantially increase the effectiveness of the distribution process. Such an implementation is currently under investigation.

\end{abstract}
\section{Introduction}

Over the past few years, the value of global e-commerce sales has been steadily increasing, leading to a considerable rise in the number of parcels being shipped worldwide every day \cite{covid_boom}. The effective delivery of parcels relies on the crucial role played by delivery companies and post offices in connecting senders with recipients. Therefore, it is essential that these companies have efficient methods to ensure successful deliveries of parcels. Although most parcels have accurate address information, there are instances where addresses are written in an unstructured way, leading to incorrect or failed deliveries. The errors may include insufficient information, redundant information, or spelling mistakes, among others.
\par
While there is no publicly available information on how companies address these issues, some methods involve address normalization, such as converting “Street” to “St.”, or parsing the address elements followed by pair-wise matching. However, these techniques are not perfect and frequently require human intervention.
\par
The primary objective of this work is to develop a solution that can enhance the quality of postal and parcel delivery services by reducing the number of misdelivered parcels and minimizing human involvement. Given the recent advancements that transformers have provided in the natural language processing field, we have chosen a fully transformer-based architecture for our solution. We combine a siamese neural network (bi-encoder: retriever) with a DistilBERT \cite{DistilBERT} model adapted for sentence-pair classification (cross-encoder: reranker). To the best of our knowledge, our work is the first one to use this type of approach to tackle an address-matching task.

\section{Background and Related Work}
\label{RelatedWork}
Determining if two addresses refer to the same location can be a challenging task. The most straightforward method for this, is to calculate the similarity metrics between the strings that describe each address. The standard algorithm used for this purpose is the edit distance, also known as the Levenshtein distance (LD) \cite{LevDistance}. However, the LD fails to provide accurate results, even for simple cases. To address this issue, more sophisticated algorithms have been developed, such as searching for the largest sub-sequences of common words, or tokenizing strings by words and sorting them alphabetically, so that the original word order is not relevant \cite{FuzzyWuzzy}. Nevertheless, the effectiveness of string similarity measures for string matching varies depending on the task, and no single algorithm can be claimed to be superior \cite{ToponymMatching}. 
\par
A significant challenge associated with traditional string similarity measures is to choose an appropriate threshold that determines when a match is considered correct or not. Santos et al. addressed this issue by proposing a supervised machine learning approach that leverages string similarity values as the model features \cite{BrunoToponym}. This method reduced the need for manual threshold tuning and improved the matching performance against the more traditional approaches.
\par
The methods previously discussed are effective at identifying symbolic similarities between addresses. However, they often struggle to accurately match addresses that share semantic meaning but are written differently. Deep Learning (DL) techniques have brought a new level of flexibility to string matching algorithms by leveraging sentence-level features that capture semantic similarities. As a result, recent studies in address matching have shifted towards DL methods due to their ability to produce superior results \cite{DeepSemanticAddress}.
\par
Comber et al. proposed a novel approach \cite{CRF+Word2Vec} that leverages the benefits of both conditional random fields (CRFs) and Word2Vec \cite{Word2Vec} for address matching. CRFs are employed to parse the address into its main components, and then Word2Vec is used to create an embedding for each parsed field. 
The similarity between fields is computed using cosine similarity, and a machine learning classifier is then used to determine whether the two addresses match. The author's proposed approach outperformed previous techniques such as CRF + Jaro-Winkler similarity \cite{CRF+Word2Vec}.
\par

Lin et al. proposed solving an address matching task using the Enhanced Sequential Inference Model (ESIM). The first step is to train a Word2Vec model to transform address records into their corresponding vector representations. Then, the ESIM model is applied, which consists of four main steps. Firstly, the input addresses are encoded using a Bi-LSTM. Then, local inference is performed on the encoded addresses through a decomposable attention mechanism. Next, a new bidirectional long short-term memory (Bi-LSTM) layer is applied to extract higher-level representations of the addresses. Finally, a multilayer perceptron (MLP) is used to indicate whether the address pairs are a match. The proposed Word2Vec + ESIM approach outperformed simpler methods, such as Word2Vec + Random Forest, demonstrating its effectiveness for address matching \cite{DeepSemanticAddress}.
\par
Another alternative solution that has been proposed to address the issue of address matching is the Attention-Bi-LSTM-CNN (ABLC) network based on contrast learning, which has demonstrated better performance than the ESIM model \cite{ABLC}. The ABLC model combines an attention mechanism, Bi-LSTMs, and convolutional neural networks (CNN) to extract features from the addresses.
\par
A distinct methodology that has been proposed for address matching and is also relevant for multiple similarity search tasks involves the use of the best match 25 (BM25) algorithm in conjunction with BERT \cite{BERT+BM25}. The method starts by employing the BM25 algorithm to retrieve the top-10 most probable records from a database for a given query. BERT \cite{OriginalBERT} is then applied to rerank the retrieved candidates. This approach has demonstrated superior performance when compared to other models, such as Word2Vec + ML Classifiers.
\par
For similarity search tasks, pre-trained deep transformers have proven to be highly effective \cite{OriginalBERT}. There are two main types of transformers that are commonly used: cross-encoders \cite{CrossEncoderPaper1}, \cite{CrossEncoderPaper2}, \cite{CrossEncoderPaper3}, which use full self-attention to encode the pair, and dual-encoders, which encode the pair separately. Dense Passage Retrieval (DPR) \cite{DensePassageRetrieval} and SBERT \cite{SBERT} are two well-established dual-encoder approaches widely used for similarity search tasks.

\section{Data Description and Preparation}
\subsection{Addresses Structure}

\begin{figure}
 \centering \includegraphics[width=0.60\textwidth]{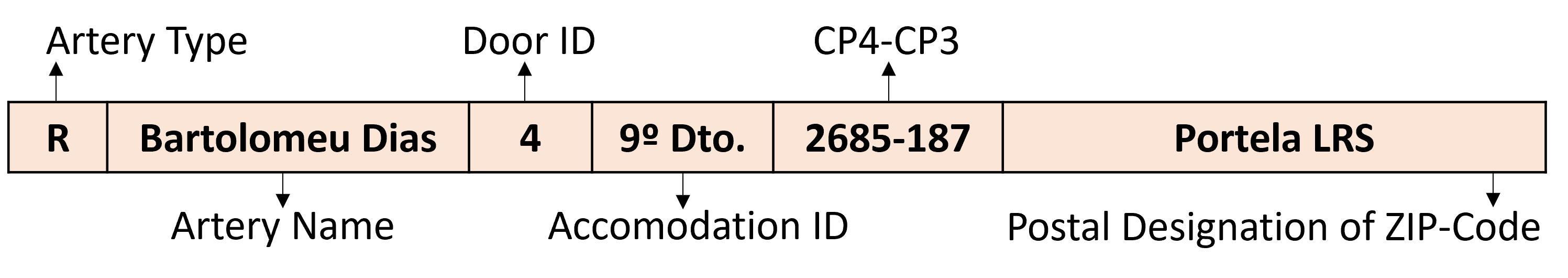}
 \caption{Example of a normalized portuguese address}
 \label{fig:ExNormAddress}
\end{figure}

This work employed two main types of addresses: (i) normalized addresses - follow a specific structure and adhere to predefined rules and (ii) unnormalized addresses - often unstructured and, therefore, more difficult to interpret. When sending a parcel, the sender usually writes the recipient's address in an unnormalized format. Classifying an address as unnormalized does not necessarily mean that it is incorrect, but rather that it does not fully comply with the standardized structure. A typical normalized Portuguese address comprises several essential elements, including (1) Artery Type - the configuration of the artery; (2) Artery Name; (3) Door ID - the house or apartment number; (4) Accommodation ID - details about the floor and accommodation and (5) ZIP-Code - a 7-digit code followed by a Postal Designation determined by the Post Office (known as CP4-CP3 combination). Figure \ref{fig:ExNormAddress} provides an example of a normalized Portuguese address.

\subsection{Datasets}
\subsubsection{Normalized Dataset}

The normalized addresses dataset used in this work was made available by CTT-Correios de Portugal, the national post office company of Portugal. The dataset comprises approximately 430k addresses, which corresponds to roughly 10\% of the universe of addresses in Portugal. Although not all addresses are included, the provided data covers the entire country and not just a specific region. A detailed geographical distribution of the available addresses can be found in Appendix \hyperref[sec:Appendix B]{A}. As the data was previously curated by CTT, no cleaning steps were required.

\subsubsection{Dataset for fine-tuning the Bi-Encoder}

The dataset used to fine-tune the bi-encoder model consists of pairs of unnormalized-normalized addresses, along with a label indicating whether they match. The unnormalized address data was also obtained from CTT, based on the history of delivered parcels over a 3-month period, which resulted in over 3 million records. However, the data required deduplication and cleaning. The deduplication process consisted in removing exact duplicates, while the cleaning process restructured some records with information in the wrong columns and discarded others that lacked mappings to the normalized database. The resulting cleaned unnormalized addresses file contained approximately 1.1 million valid records.
\par
For fine-tuning the bi-encoder, 90\% of these records were sampled and duplicated to form address pairs with a 1:1 positive-to-negative ratio. The normalized address for the false matching pair is generated from three categories, each with an equal probability of occurrence: easy match (random address), hard match (address with a string similarity metric $>$ 0.8), and very hard match (address in the same ZIP-Code). This approach was chosen to increase the number of challenging records in the training dataset.

\subsubsection{Test Dataset and Dataset for fine-tuning the Cross-Encoder}
Approximately 120k unnormalized records were not used for fine-tune the bi-encoder. From this pool of records, we extracted a random sample of around 60k addresses to build the test dataset used to assess the final model's performance.
\par
The remaining available addresses were used to fine-tune the cross-encoder, with an approximate positive-to-negative ratio of 1:9. Negative samples were generated by querying the bi-encoder with the unnormalized address and retaining the top-9 most probable addresses that did not match with the unnormalized one.

\section{Model Implementation}

The proposed model consists in combining a bi-encoder with a cross-encoder. The following subsections describe in detail these two network types and how they connect with each other in order to create the final solution.

\subsection{Bi-Encoder}

Our bi-encoder is a dual-encoder network trained in a siamese way with the purpose of learning how to derive meaningful sentence embeddings that can be compared with others through cosine-similarity.
\par
\begin{figure}
  \centering \includegraphics[width=0.45\textwidth]{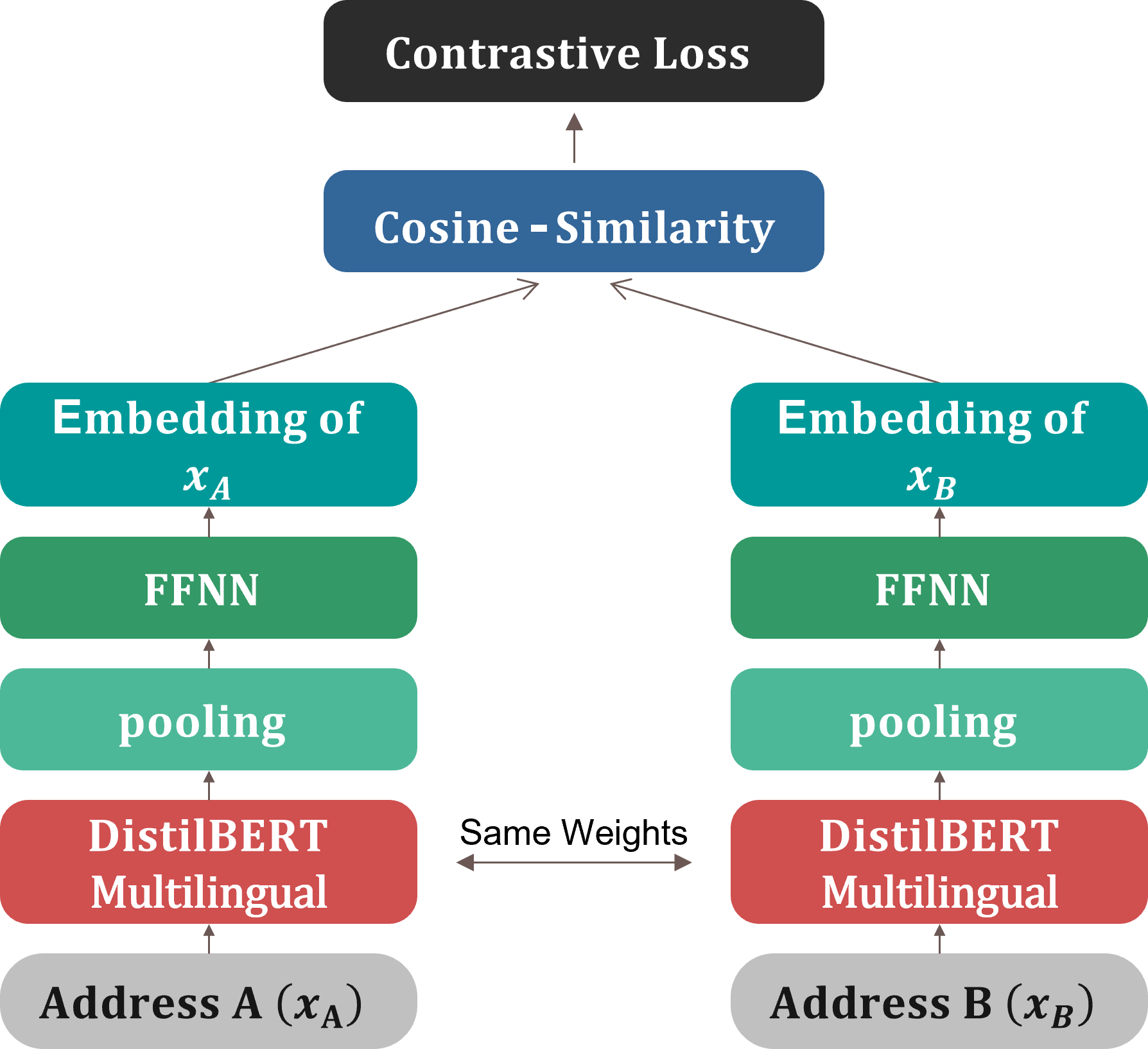}
  \caption{Bi-encoder architecture considered for fine-tuning on the address data}
  \label{fig:bi-architecture}
\end{figure}

The base transformer model used is the multilingual DistilBERT - a smaller, cheaper, and lighter version of the multilingual BERT. A distilled model is achieved through a process called knowledge distillation \cite{Distilling}, which consists in compressing a bigger model (the teacher) into a more compact model (the student) that is trained to reproduce the behavior of the teacher. It is proved that on inference time, the distilled model can be 60\% faster than the teacher, while 40\% smaller and retaining 97\% of the performance \cite{DistilBERT}. Moreover, from a deployment perspective, the adoption of a more compact model is generally preferred. Therefore, the tradeoff offered by DistilBERT was considered good enough to be chosen for this work. Even though DistilBERT was already pre-trained in some natural language processing (NLP) tasks, in order for its parameters to reach an optimal value for the specific address matching task, the model must be fine-tuned. The architecture of the bi-encoder considered for fine-tuning on the address data is displayed in Figure \ref{fig:bi-architecture}. 

\par
To generate fixed-length embeddings of the addresses, we apply mean pooling on the DistilBERT output, which is then passed through an MLP with hyperbolic tangent activation. This reduces the dimensionality of the address embeddings to 512. We employ the contrastive loss function as our optimization objective, which seeks to minimize the distance between the embeddings of matching address pairs and maximize the distance between non-matching pairs.

\begin{equation}
  \frac{1}{2}[y\cdot D^2(x_A , x_B) + (1-y) \cdot \{\textrm{relu}(\alpha - D(x_A , x_B))\}^2]
\end{equation}

\noindent
Here, $x_A$ and $x_B$ represent the embeddings of addresses A and B, respectively, while $y$ is the label indicating whether both addresses are related. The distance metric $D$ between $x_A$ and $x_B$ is calculated as 1 - cosine similarity($x_A$, $x_B$). Additionally, the margin $\alpha$ is introduced to ensure that the negative pair is at least separated by a distance equal or greater than that value.

\subsection{Cross-Encoder}
\begin{figure}[H]
  \centering \includegraphics[width=0.45\textwidth]{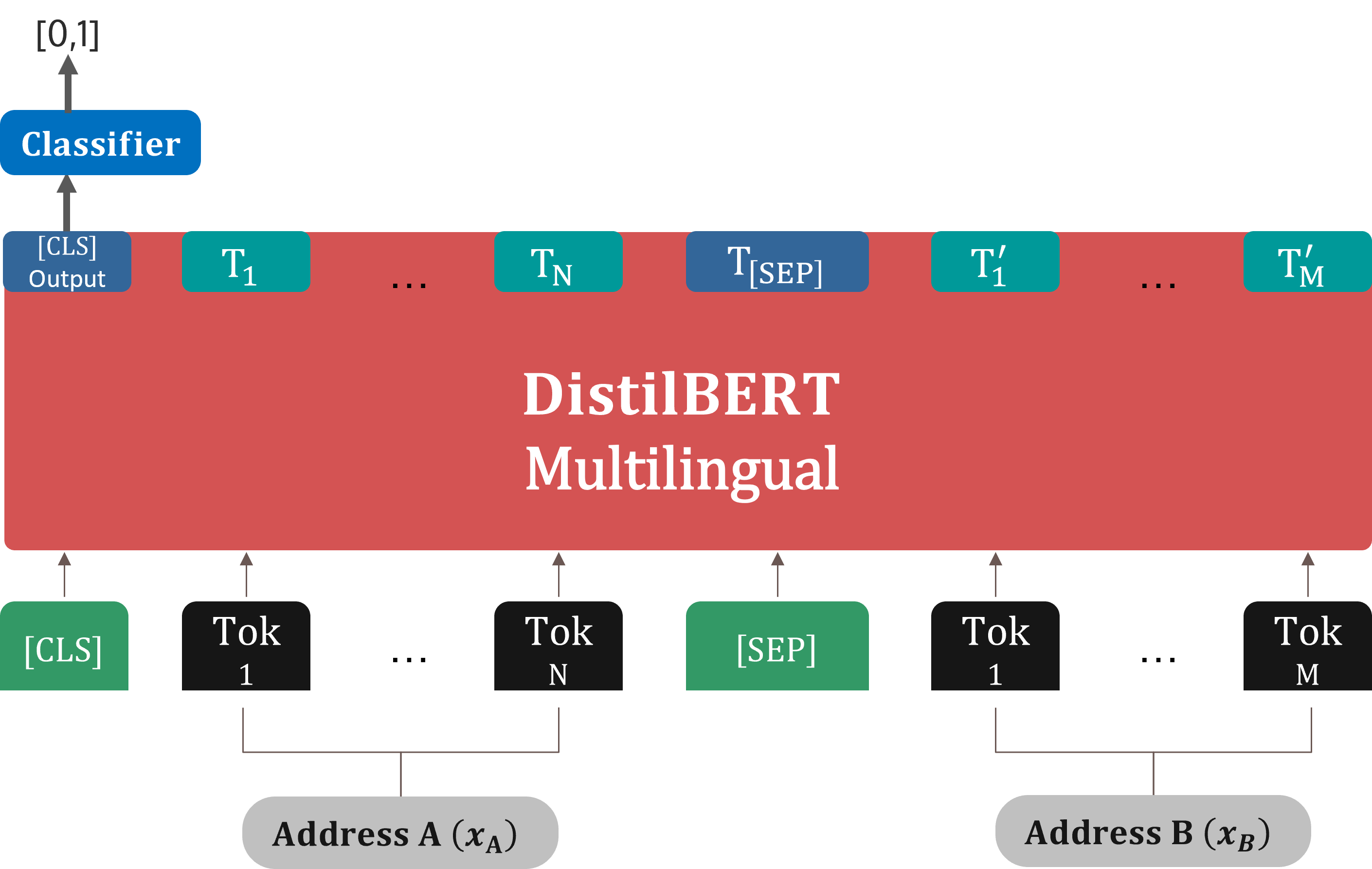}
  \caption{Cross-encoder architecture for an address matching classification task (adapted from \cite{OriginalBERT} and \cite{SBERT})}
  \label{fig:cross-architecture}
\end{figure}

One of the specificities of a bi-encoder is that sentences are given individually to the network, for which individual sentence embeddings are then computed, that can afterwards be compared through a similarity measure. A cross-encoder does the exact opposite - it feeds both sentences simultaneously to the network, like in the BERT architecture adapted for a sentence pair classification task (Figure \ref{fig:cross-architecture}). For that reason, a cross-encoder does not compute sentence embeddings.

\subsection{Proposed Model: Bi-Encoder + Cross-Encoder}

\begin{figure}
  \centering \includegraphics[width=0.8\textwidth]{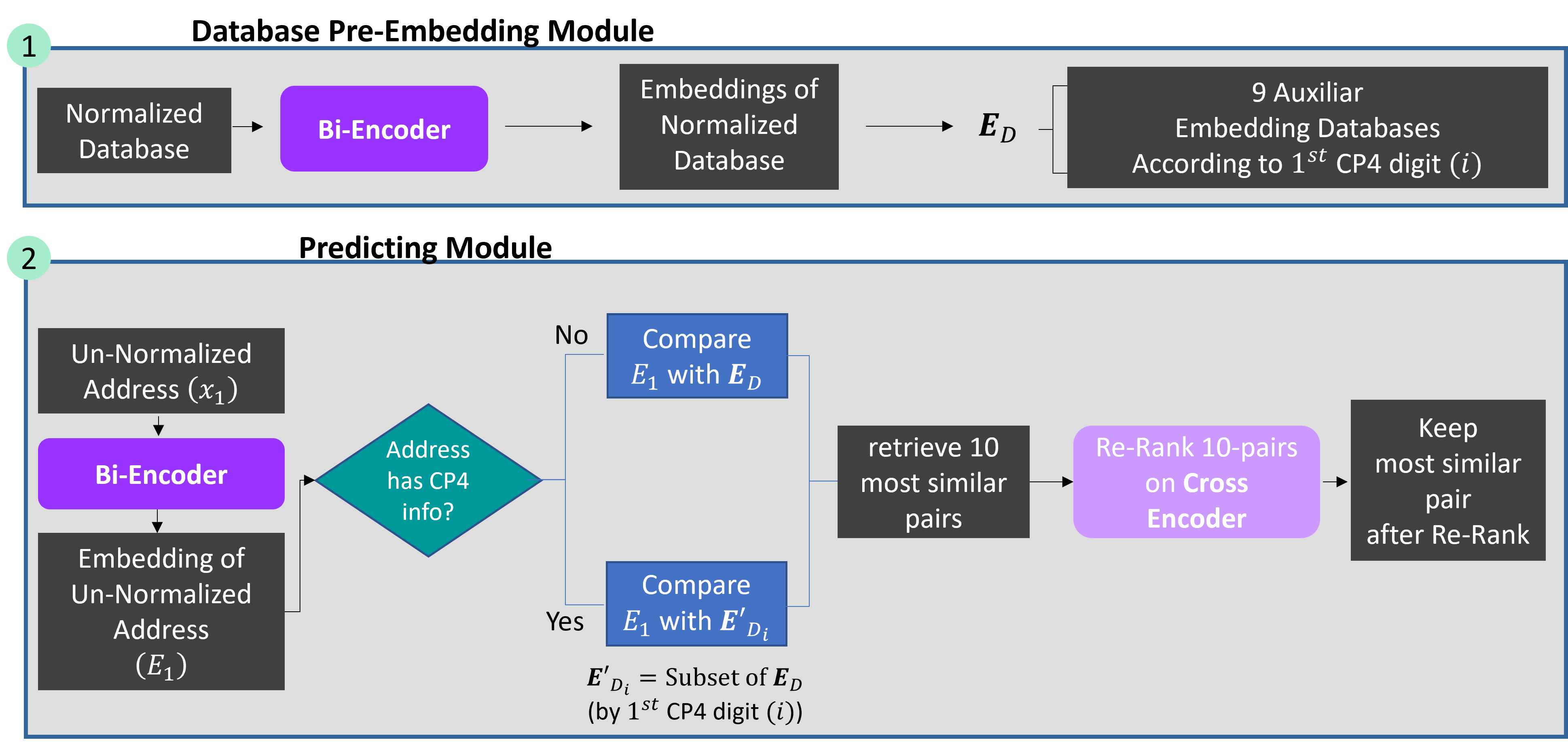}
  \caption{Architecture of the bi-encoder + cross-encoder model}
  \label{fig:ProposedModel}
\end{figure}

Reimers et al. noted that performance-wise, for a sentence similarity task, the cross-encoder achieves a better performance than a bi-encoder \cite{SBERT}. However, using only the cross-encoder for address matching is not feasible. If one wants to search on a normalized database for the address most similar to the unnormalized address that is being paired, all the combinations of (unnormalized, normalized$_i$) must be fed to the cross-encoder, which is computationally demanding. For that reason, a decision was made to not use the cross-encoder by itself for the final solution, but rather a combination of the cross-encoder with the bi-encoder, in order to get the best features of each model. The proposed model is named Bi-Encoder + Cross-Encoder or BI+CE. In Figure \ref{fig:ProposedModel} is presented the full architecture of the model for the address matching task. There are two main modules in our architecture: (1) the database pre-embedding module and (2) the predicting module.

\par
The database pre-embedding module is not mandatory but, if included, increases the speed of the predicting module significantly. Its goal is to create and store in memory all the embeddings related to the normalized database. Since the database will be static most of the time, it is unnecessary to recompute embeddings every time the model is initialized. The module receives as input each normalized address and computes its corresponding embedding through the bi-encoder. The final outputs are aggregated on a normalized embeddings file. In order to increase the performance speed in the predicting module, an extra step is performed: nine auxiliar databases are created according to the nine possible first digits in a CP4. Each unnormalized address is, therefore, on the predicting stage, compared only with the addresses on the corresponding auxiliar database.
\par
The predicting module is the main part of the BI+CE model. The process of finding the corresponding pair in the normalized database for the target unnormalized address ($x_1$)  is done by: (i) feeding $x_1$ to the bi-encoder; (ii) comparing the embedding of $x_1$ ($E_1$) with the correspondent auxiliar embeddings through cosine-similarity; (iii) returning the k-most similar addresses (in this case k=10); (iv) feeding the pairs ($x_1$,returned$_{address_i}$) to the cross-encoder, which will rerank them.
\par
The reason behind selecting the top-10 addresses instead of only the most similar is due to the fact that the bi-encoder sometimes misses at assigning the highest probability to the correct address. However, the correct address is usually retrieved in the top-10, hence, the cross-encoder is used. More details on this topic are provided in section \ref{RESULTS}.

\subsection{Training Overview}

Table \ref{tab:hyperparameters} displays the combination of  the best hyperparameters, chosen for fine-tuning the bi-encoder and the cross-encoder. They were reached by trial and error using the common good practices to fine-tune this type of models \cite{SBERT}.
\par
Regarding the values for the bi-encoder, the variable that is further apart from the usual values is the epoch number. Usually this variable is never higher than 4. However, we decided to select a value of 20. A study was performed on the impact of the epoch number on the fine-tuning performance of models like BERT, and the conclusions were that a larger number of epochs, such as 20, works better \cite{FineTuneBERT}. As for the cross-encoder, we explored various epoch values, including 20, in order to keep the hyperparameters similar to the bi-encoder, but determined that 15 yielded the best results. In Appendix \hyperref[sec:Appendix G]{B} and Appendix \hyperref[sec:Appendix H]{C} are presented the hyperparameter combinations that were tested in order to reach the optimal values for maximizing the performance of the bi-encoder and the cross-encoder on the test dataset.
\par
Both bi-encoder and cross-encoder are fine-tuned using one NVIDIA Tesla V100S (32 GB)\footnote{Code available at: \url{https://github.com/avduarte333/adress-matching}}. Each fine-tuning process takes approximately eight hours to complete and is done using the python package ‘sentence-transformers’\footnote{\url{https://github.com/UKPLab/sentence-transformers}}. The experimental results for the traditional approaches consist of individual runs for each model. However, we conducted multiple runs of both the bi-encoder and the cross-encoder models after obtaining the optimized hyperparameters. To ensure the robustness of our findings, we run each model 5 times and report the results using the ones with the most consistent performance across the runs, as determined by the median outcome accuracy at the door level. Our results demonstrate that fine-tuning remains stable across the runs. We observed a low standard deviation of 0.077 for door level accuracy in the bi-encoder and 0.084 in the cross-encoder (more details about the 5 runs in Appendix \hyperref[sec:Appendix I]{D}).

\begin{table}
\centering
\setlength\extrarowheight{2pt}
\caption{Best found fine-tuning parameters for  bi-encoder and cross-encoder.}
\label{tab:hyperparameters}
\begin{tabular}{|c|cc|}
\hline
{\textbf{Parameter}} & \multicolumn{1}{c|}{{\textbf{Bi-Encoder}}}                                               & {\textbf{Cross-Encoder}} \\ \hline
Epochs                   & \multicolumn{1}{c|}{20}                                                                      & 15                           \\ \hline
Batch Size                 & \multicolumn{2}{c|}{16}                                                                                                  \\ \hline
Optmizer                 & \multicolumn{2}{c|}{AdamW}                                                                                                  \\ \hline
Learning Rate             & \multicolumn{2}{c|}{$10^{-5}$}                                                                                                   \\ \hline
Scheduler             & \multicolumn{2}{c|}{Linear}                                                                                                   \\ \hline
Warmup Steps             & \multicolumn{2}{c|}{100}                                                                                                   \\ \hline
Weight Decay             & \multicolumn{2}{c|}{0.01}                                                                                                   \\ \hline
Loss                     & \multicolumn{1}{c|}{\begin{tabular}[c]{@{}c@{}}Contrastive Loss\\ margin = 0.5\end{tabular}} & Cross-Entropy                \\ \hline
Base Transformer         & \multicolumn{2}{c|}{Multilingual DistilBERT}                                                                                             \\ \hline
\end{tabular}
\end{table}

\subsection{Model Evaluation}

For comparison purposes, we evaluated several models: (i) the proposed one (BI+CE), (ii) two traditional string matching algorithms, token sort and token set \footnote{\url{https://github.com/seatgeek/fuzzywuzzy}}, (iii) a bi-encoder, (iv) a BM25 ranking function combined with a cross-encoder (BM25+CE), based on the approach of Gupta et al. \cite{BERT+BM25}, and (v) a Dense Passage Retrieval (DPR) model as introduced by Karpukhin et al. \cite{DensePassageRetrieval}, where we used two independent pre-trained multilingual DistilBERTs as base transformers. 
\par
The models were evaluated based on two metrics: (1) inference time, which is the number of matches performed by the model in one second, and (2) accuracy, which is the proportion of correctly predicted pairs out of all pairs. Additionally, we analyzed the quality of the top-k retrieval for the approaches (iii), (iv), and (v).

\par
The primary objective of the model is to achieve high accuracy at the door level, as correctly identifying the door is crucial for parcel delivery services. While retrieving the correct artery is important, failure to identify the correct door results in an incorrect mapping. However, misidentifying doors in the same artery should not significantly impact delivery efficiency, as they are typically close in geographic proximity. Thus, the results are reported for both artery and door level accuracy.

\par
In practical applications, it is crucial to minimize the number of misdelivered parcels. To achieve this, we propose imposing a threshold or filter value (cutting value) on the matching probability variable to ensure that only address pairs with high matching probability are accepted as correct, and the remaining pairs are subjected to manual inspection. In all experiments, we used this criterion and selected the optimal filter values by examining the match confidence variable's distribution.

\section{Results and Discussion}
\label{RESULTS}
\subsection{Inference Time}

The average number of unnormalized addresses paired per second for each tested model is presented in Table \ref{tab:ItPerSecond}. To ensure a fair comparison, we made sure that each model performs roughly the same number of operations when pairing new addresses.

\begin{table}[H]
\centering
\caption{Average number of iterations / second for each model (it/s). We run the DL models on GPU. However, instead of batching, we process each address individually, as displayed in Figure \ref{fig:ProposedModel}, to perform a fairer comparison to the traditional approaches. The best results are highlighted in bold.}
\label{tab:ItPerSecond}
\setlength\extrarowheight{2pt}
\begin{tabular}{c|c|c|c|c|c|c|}
\cline{2-7}
 & Token Sort & Token Set & BM25+CE & DPR & Bi-Encoder & BI+CE \\ \hline
\multicolumn{1}{|c|}{Without CP4 Filter} & 0.01 & 0.07 & 0.09 & 0.76 & \textbf{0.91} & 0.80 \\ \hline
\multicolumn{1}{|c|}{With CP4 Filter} & 0.96 & 0.61 & 0.83 & 3.64 & \textbf{5.40} & 3.70 \\ \hline
\end{tabular}
\end{table}

\par
As anticipated, the inference speed improves when CP4 filtering is applied, regardless of the approach. Without the CP4 filter, the inference time ranges from 0.07 to 0.91 iterations per second, while with CP4 filter, the inference time ranges from 0.61 to 5.40 iterations per second. However, there is a significant difference in the performance between the traditional string matching algorithms and the ones where dense vectors are utilized (DPR, bi-encoder and BI+CE). This can be attributed to the fact that calculating cosine similarity between an address and the candidate addresses in the normalized database is a faster operation than computing string metrics.
\par
The results presented in Table \ref{tab:ItPerSecond} also reveal that adding a cross-encoder to the model architecture decreases the model's inference speed. This is expected, as the extra layer of complexity introduced by the cross-encoder leads to an increase in computational workload.

\subsection{Accuracy Results - Test Dataset}

\subsubsection{Results - Traditional String Matching Algorithms}

From Table \ref{tab:all_results}, both results at artery and door level suggest that the token set algorithm performs better than the token sort, since it usually achieves higher accuracies while retaining more addresses (27.09\% vs 17.57\% at door level). However, when performing the manual filtering, the great majority of addresses are discarded, and there is no significant improvement in the overall accuracy. Results at artery level are significantly better than the ones at the door level but none of the algorithms achieved results that may be considered promising enough to solve the address matching problem successfully.

\subsubsection{Results - Bi-Encoder, DPR and BM25+CE}

When evaluating the retrieval capabilities of the bi-encoder, DPR and the BM25+CE\footnote{Although the model under study is the BM25+CE, when evaluating the retrieval capabilities, the cross-encoder is not used, therefore, for notation simplicity, the model is mentioned as BM25.}, one can consider two scenarios: the top-1 retrieval and the top-k. Table \ref{tab:top1vs10} displays, for each method, the proportion of instances where the correct normalized address is among the retrieved addresses.

\begin{table}[H]
\centering
   \caption{Accuracy (\%) at artery and door level for the tested approaches as a function of the filter value. The best results are highlighted in bold.}
    \label{tab:all_results}
\setlength\extrarowheight{2pt}
\begin{threeparttable}[b]
   \centering
\begin{tabular}{|c|c|c|c|c|c|c|}
\hline
Algorithm & \textbf{\begin{tabular}[c]{@{}c@{}}No Filter\\ Artery\end{tabular}} & \textbf{\begin{tabular}[c]{@{}c@{}}No Filter\\ Door\end{tabular}} & \textbf{\begin{tabular}[c]{@{}c@{}}Filter\\ Value\end{tabular}} & \textbf{\begin{tabular}[c]{@{}c@{}}With Filter\\ Artery\end{tabular}} & \textbf{\begin{tabular}[c]{@{}c@{}}With Filter\\ Door\end{tabular}} & \textbf{\begin{tabular}[c]{@{}c@{}}Discarded\\ Addresses\end{tabular}} \\ \hline
Token Sort & 51.09 & 17.57 & 0.85 & 68.40 & 25.60 & 92.80 \\ \hline
Token Set & 64.67 & 27.09 & 0.85 & 69.44 & 32.20 & 54.80 \\ \hline
BM25 + CE & 79.78 & 63.31 & 0.95 & 98.47 & 84.34 & 38.03 \\ \hline
DPR & 95.48 & 85.91 & 0.90 & 99.50 & 91.50 & 19.80 \\ \hline
Bi-Encoder & 96.49 & 95.68 & 0.99 & 97.82 & 97.39 & 23.37 \\ \hline
BI+CE & \textbf{97.08} & \textbf{95.32} & 0.90 & \textbf{99.71} & \textbf{98.35} & \textbf{18.08} \\ \hline
\end{tabular}
  \end{threeparttable}
\end{table}

\begin{table}[H]
\centering
\caption{Proportion of instances (\%) where the correct normalized address (door level) is among the retrieved addresses. The best results are highlighted in bold.}
\label{tab:top1vs10}
\setlength\extrarowheight{2pt}
\begin{tabular}{c|c|c|c|}
\cline{2-4}
                             & \; \; \textbf{Bi-Encoder} \; \; & \;\textbf{DPR(no rerank)}\; & \textbf{\; \; \; \;BM25\; \; \; \;} \\ \hline
\multicolumn{1}{|c|}{\; \; Top-1\; \; }  & \textbf{95.68}             & 80.92                        & 33.49       \\ \hline
\multicolumn{1}{|c|}{Top-10} & \textbf{99.41}            & 97.02                        & 72.80       \\ \hline
\end{tabular}
\end{table}

It is evident from Table \ref{tab:all_results} and Table \ref{tab:top1vs10} that introducing the dense retrievers, like the bi-encoder or DPR, in the solution enhances the results significantly. Its accuracies both on artery and door level are above 85\%, while for BM25 and the traditional methods they never surpass 63.31\%. 
\par

Table \ref{tab:top1vs10} also highlights two major advantages of the bi-encoder in terms of retrieval quality. Firstly, the top-1 retrieval alone is a near-perfect solution, with a door level accuracy of 95.68\%. Secondly, the top-10 retrieval gives almost every address a chance of being correctly paired. Our experimental results indicate that while DPR achieves a top-10 accuracy comparable to that of the bi-encoder, its top-1 accuracy falls short by nearly 15\% (95.68\% - 80.92\%). Regarding BM25, its top-1 accuracy is limited to 33.49\% and 72.80\% at most when considering top-10 retrieval. Despite being more than double, it still falls short compared to the bi-encoder. Therefore, using the BM25 or DPR as a solution for the problem is not optimal. Nevertheless, it is worth mentioning that the rerankers can leverage the retrieval results of the BM25 and the DPR significantly. When considering the top-1 address before and after the reranking, the door level accuracies shift from 33.49\% to 63.31\% for BM25 and from 80.92\% to  85.91\% for DPR (Table \ref{tab:all_results}).

\par
\begin{figure}%
    \centering
    \subfloat[\centering Bi-Encoder]{{\hspace{0.2cm}\includegraphics[width=6.2cm]{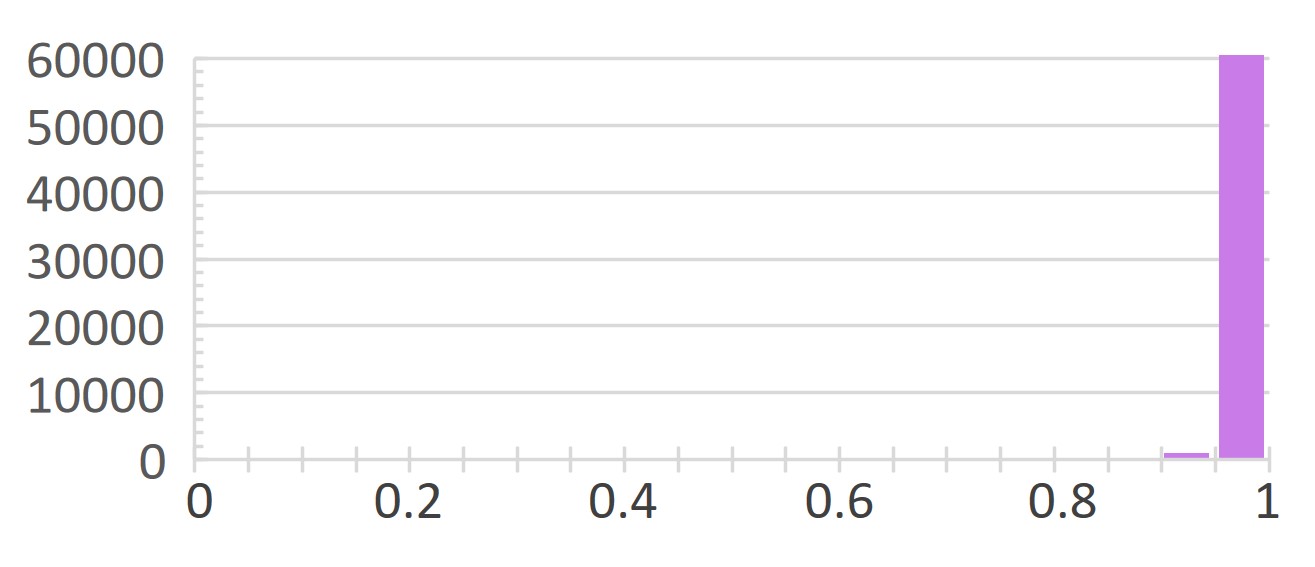} }
    \label{fig:BiHistogram}}%
    \qquad
    \subfloat[\centering Bi+CE]{{\hspace{0.25cm}\includegraphics[width=6.2cm]{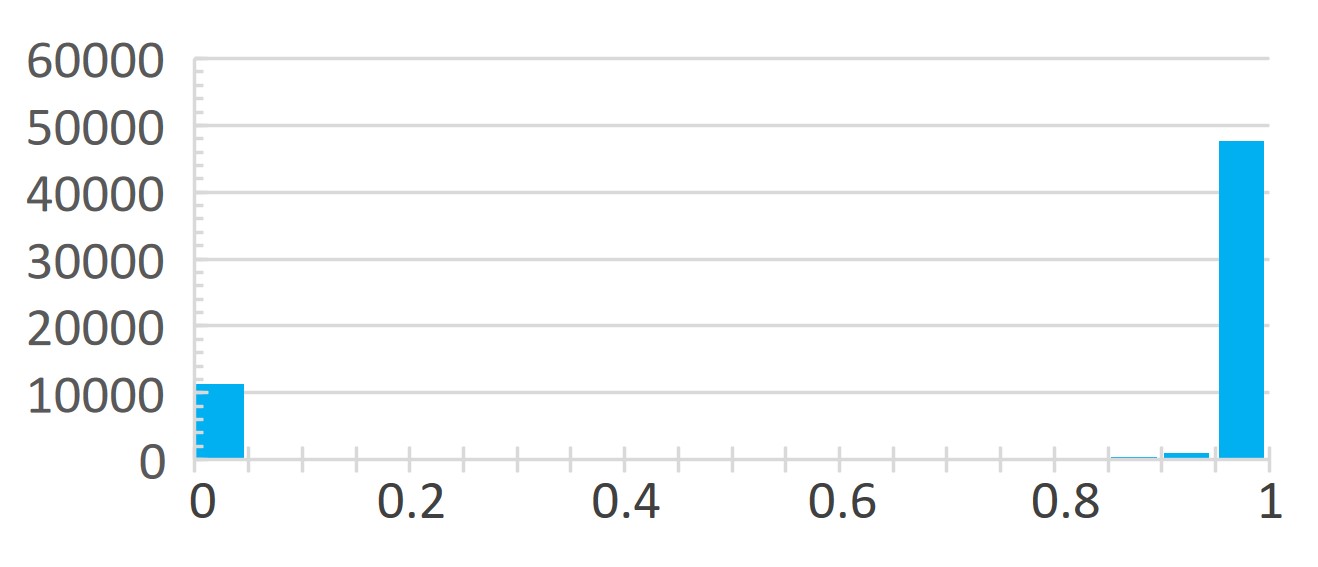} }
    \label{fig:BiCEHistogram}}%
    \caption{Histogram of the matching probability assigned by the bi-encoder model (left) and the BI+CE model (right).}%
    \label{fig:example}%
\end{figure}

When studying the optimal cutting value for filtering the bi-encoder results, two interesting properties in the distribution of the match confidence variable (Figure \ref{fig:BiHistogram}) were identified: (i) distribution strongly skewed and (ii) ~77\% of the pairs have a matching probability that lies in the [0.99;1.00] interval. Combining these factors, the cutting value chosen was 0.99.

\par
The hypothesis that implementing this filtering technique would result in near-perfect classification was not supported by the data (Table \ref{tab:all_results}). Despite an overall increase in the model's accuracy (95.68\% to 97.39\% at door level) and the fact that only 23.37\% of records were discarded, there remains room for improvement. Therefore, in light of these findings and the ones from the BM25 and DPR experiments, it was decided to incorporate the cross-encoder into the proposed model.

\subsubsection{Results - BI+CE (Proposed Model)}

Contrary to the expectation, the model’s overall accuracy at the door level did not improve in comparison to the bi-encoder approach – 4.68\% of the addresses remain incorrectly classified (Table \ref{tab:all_results}: 100\% - 95.32\%). It improved, though, at the artery level, but only slightly (96.49\% to 97.08\%). The matching probability distribution is, however, quite different in this scenario (Figure \ref{fig:BiCEHistogram}).

\par
In the bi-encoder experiment, the lowest matching probability assigned by the model was 0.689. In the  BI+CE model, the lowest probabilities assigned are really low values ($<$ 1\%). Figure \ref{fig:BiCEHistogram} displays a big gap between the highest probable pairs and the lowest probable ones. There are a few pairs spread across the x-axis scale. However, their proportion is just 2.28 \% of the total number of addresses. The cutting value chosen for filtering, in this case, is 0.90. Performing this step provided interesting results, namely: the variation of the accuracy on artery and door level is quite positive (from 97.08\% to 99.71\% on artery and from 95.32\% to 98.35\% on door) and the number of discarded addresses is lower than the number discarded on the bi-encoder experiment (from 23.37\% to 18.08\%).

\section{Conclusions}

The main goal of this work was the development of a model that could solve with success an address matching task, by using a DL approach, specifically pre-trained transformers. The bi-encoder proved to be a fundamental piece in the solution, not only for the speed up it introduces but also for its retrieval quality which can place the correct normalized address in the top-10 retrievals 99.41\% of the time. We also found that the cross-encoder increases the robustness of the model’s accuracy, at the cost of a negative impact on the inference time. Nevertheless, that drawback can be mitigated by using the model with GPU computations where the inference speed can significantly increase against more traditional approaches such as the BM25 (roughly 4.5 times faster).
\par
In a real application, we would probably assume that the only correct pairs are the ones that the model gave a high matching probability ($>$ 0.90). The results in the test dataset suggest that imposing such criteria would significantly reduce the number of misdelivered packages, although a small proportion of the addresses ($\sim$18\%) would still require a manual correction. There are other alternatives, such as disregarding the matching probability variable, which would mitigate the time spent on manual correction. It would, however, introduce some downsides such as a higher error rate on the package delivery.

\section{Acknowledgements}
The authors would like to acknowledge the support of Dr. Egídio Moutinho, Drª. Marília Rosado, Dr. Rúben Rocha, Dr. André Esteves, Dr. Paulo Silva, Dr. Gonçalo Ribeiro Enes and Dr. Diogo Freitas Oliveira in the development of this project. We also gratefully acknowledge the financial support provided by Recovery and Resilience Fund towards the Center for Responsible AI project (Ref. C628696807-00454142) and the multiannual financing of the Foundation for Science and Technology (FCT) for INESC-ID (Ref. UIDB/50021/2020).

\printbibliography
\newpage

\appendix

\section{Distribution of addresses per region on normalized database}
\label{sec:Appendix B}

\begin{figure}[H]
  \centering \includegraphics[width=0.95\textwidth]{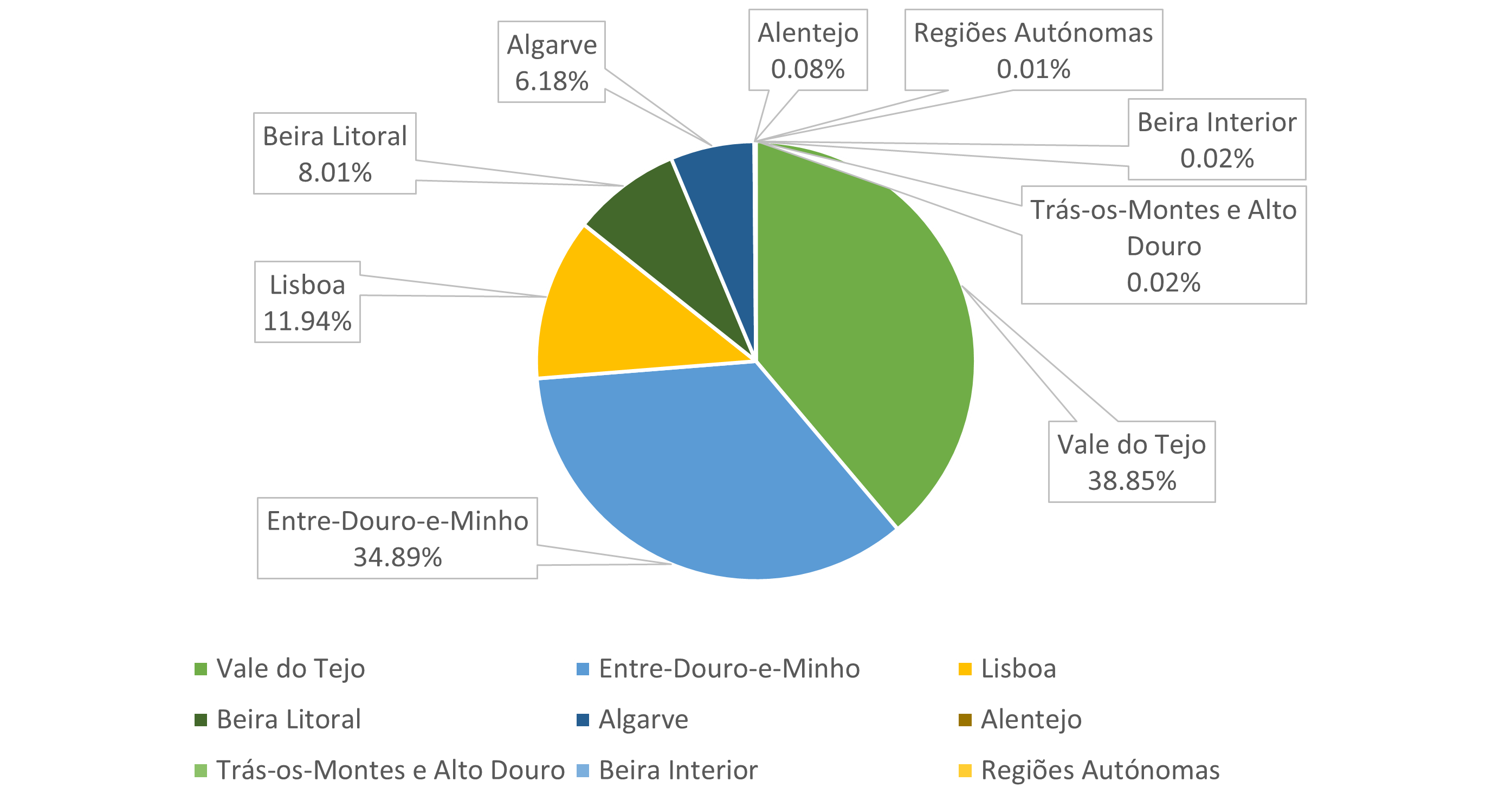}
  \caption{Distribution of Addresses Per Region on Normalized Database}
  \label{fig:AddressDistribution}
\end{figure}

\section{Best hyperparameter search for Bi-Encoder fine-tuning}
\label{sec:Appendix G}

\begin{table}[H]
\centering
  \caption{Search for the hyperparameters that maximize the door level accuracy (\%) of the bi-encoder on the test dataset. The selected model is highlighted in the purple cells.}
    \label{tab:bi_hiperparameter_search}
\setlength\extrarowheight{2pt}
\begin{threeparttable}[b]
   \centering
\begin{tabular}{|c|c|c|c|c|c|}
\hline
\textbf{\begin{tabular}[c]{@{}c@{}}Base\\ Transformer\end{tabular}} & \; \textbf{\begin{tabular}[c]{@{}c@{}}Batch\\ Size\end{tabular}}\; & \; \textbf{Epochs} \;  & \; \textbf{\begin{tabular}[c]{@{}c@{}}Contrastive\\ Loss Margin\end{tabular}\; } & \; \textbf{\begin{tabular}[c]{@{}c@{}}Learning\\ Rate\end{tabular}} \;  & \; \textbf{\begin{tabular}[c]{@{}c@{}}Door\\ Accuracy\end{tabular}} \;  \\ \hline
\rowcolor[HTML]{CBCEFB} 
DistilBERT & 16 & 20 & 0.5 & 1e-5 & 95.68 \\ \hline
DistilBERT & 16 & \textbf{6} & 0.5 & \textbf{2e-5} & 93.42 \\ \hline
DistilBERT & 16 & \textbf{6} & \textbf{0.7} & \textbf{2e-5} & 80.18 \\ \hline
DistilBERT & 16 & \textbf{5} & 0.5 & \textbf{2e-5} & 92.89 \\ \hline
DistilBERT & \textbf{32} & \textbf{7} & 0.5 & 1e-5 & 84.90 \\ \hline
DistilBERT & \textbf{32} & \textbf{1} & 0.5 & 1e-5 & 74.93 \\ \hline
DistilBERT & \textbf{32} & \textbf{5} & 0.5 & 1e-5 & 90.27 \\ \hline
DistilBERT & \textbf{32} & 20 & 0.5 & \textbf{1e-6} & 49.48 \\ \hline
XLM RoBERTa & 16 & \textbf{1} & 0.5 & \textbf{2e-5} & 83.09 \\ \hline
XLM RoBERTa & 16 & \textbf{5} & 0.5 & 1e-5 & 71.35 \\ \hline
XLM RoBERTa & \textbf{512} & \textbf{30} & 0.5 & 1e-5 & 53.58 \\ \hline
BERTimbau & 16 & \textbf{5} & 0.5 & 1e-5 & 79.12 \\ \hline
BERTimbau & 16 & \textbf{4} & 0.5 & \textbf{2e-5} & 78.48 \\ \hline
BERTimbau & 16 & \textbf{1} & 0.5 & 1e-5 & 73.99 \\ \hline
\end{tabular}
 \begin{tablenotes}
        \item [a] The parameters that were changed in each experiment in relation to the best model are identified with bold.
        \item [b] Learning Rate = AdamW, Weight Decay = 0.01, Warmup Scheduler = Linear with 100 Steps Warmup, Loss = Contrastive Loss (margin = 0.5) for all tested approaches.
 \end{tablenotes}
\end{threeparttable}
\end{table}

\section{Best hyperparameter search for Cross-Encoder Fine-Tuning}
\label{sec:Appendix H}

\begin{table}[H]
\centering
  \caption{Search for the hyperparameters that maximize the door level accuracy (\%) of the cross-encoder on the test dataset. The selected model is highlighted in the purple cells.}
    \label{tab:ce_hiperparameter_search}
    \centering
\setlength\extrarowheight{2pt}
\begin{threeparttable}[b]
   \centering
\begin{tabular}{|c|c|c|c|c|c|}
\hline
\;\textbf{\begin{tabular}[c]{@{}c@{}}Base\\ Transformer\end{tabular}}\; & \;\textbf{\begin{tabular}[c]{@{}c@{}}Batch\\ Size\end{tabular}} \;& \;\textbf{Epochs}\; & \;\textbf{\begin{tabular}[c]{@{}c@{}}Learning\\ Rate\end{tabular}}\; & \textbf{\begin{tabular}[c]{@{}c@{}}Without Filter\\ Door Accuracy\end{tabular}} & \textbf{\begin{tabular}[c]{@{}c@{}}With Filter\\ Door Accuracy\end{tabular}} \\ \hline
\rowcolor[HTML]{CBCEFB}  
DistilBERT & 16 & 15 & 1e-5 & 95.32 & 98.35 \\ \hline
DistilBERT & 16 & \textbf{20} & 1e-5 & 95.13 & 97.94 \\ \hline
DistilBERT & 16 & 15 & \textbf{2e-5} & 94.88 & 97.55 \\ \hline
\end{tabular}
 \begin{tablenotes}
        \item [a] The parameters that were changed in each experiment in relation to the best model are identified with bold.
        \item [b] Learning Rate = AdamW, Weight Decay = 0.01, Warmup Scheduler = Linear with 100 Steps Warmup, Loss = Cross-Entropy for all tested approaches.
 \end{tablenotes}
\end{threeparttable}
\end{table}

\section{Bi-Encoder and Cross-Encoder accuracy plots with error bars}
\label{sec:Appendix I}

\begin{table}[H]
\centering
  \caption{Accuracy (\%) obtained in the test dataset for the bi-encoder and the BI+CE across the 5 different runs. We select the models where the median door level accuracy is obtained (highlighted in purple).}
 \centering 
\setlength\extrarowheight{2pt}
\centering
\begin{tabular}{c|cc|cc|}
\cline{2-5}
\multicolumn{1}{l|}{} & \multicolumn{2}{c|}{\textbf{Bi-Encoder}} & \multicolumn{2}{c|}{\textbf{BI+CE}} \\ \hline
\multicolumn{1}{|c|}{\textbf{Run Number}} & \multicolumn{1}{c|}{Artery Level} & Door Level & \multicolumn{1}{c|}{Artery Level} & Door Level \\ \hline
\multicolumn{1}{|c|}{1st Run} & \multicolumn{1}{c|}{\cellcolor[HTML]{CBCEFB}96.49} & \cellcolor[HTML]{CBCEFB}95.68 & \multicolumn{1}{c|}{96.99} & 95.20 \\ \hline
\multicolumn{1}{|c|}{2nd Run} & \multicolumn{1}{c|}{96.44} & 95.50 & \multicolumn{1}{c|}{96.94} &  95.19\\ \hline
\multicolumn{1}{|c|}{3rd Run} & \multicolumn{1}{c|}{96.48} & 95.59 & \multicolumn{1}{c|}{\cellcolor[HTML]{CBCEFB}97.08} &  \cellcolor[HTML]{CBCEFB}95.32\\ \hline
\multicolumn{1}{|c|}{4th Run} & \multicolumn{1}{c|}{96.55} & 95.69 & \multicolumn{1}{c|}{97.24} &  95.41\\ \hline
\multicolumn{1}{|c|}{5th Run} & \multicolumn{1}{c|}{96.63} & 95.71 & \multicolumn{1}{c|}{97.10} & 95.33 \\ \hline
\multicolumn{1}{|c|}{Median} & \multicolumn{1}{c|}{96.49} & 95.68 & \multicolumn{1}{c|}{97.08} & 95.32 \\ \hline
\multicolumn{1}{|c|}{Standard Deviation} & \multicolumn{1}{c|}{0.066} & 0.077 & \multicolumn{1}{c|}{0.103} & 0.084 \\ \hline
\end{tabular}
\end{table}

\begin{figure}[H]
\centering
\parbox{5cm}{
\includegraphics[width=5cm]{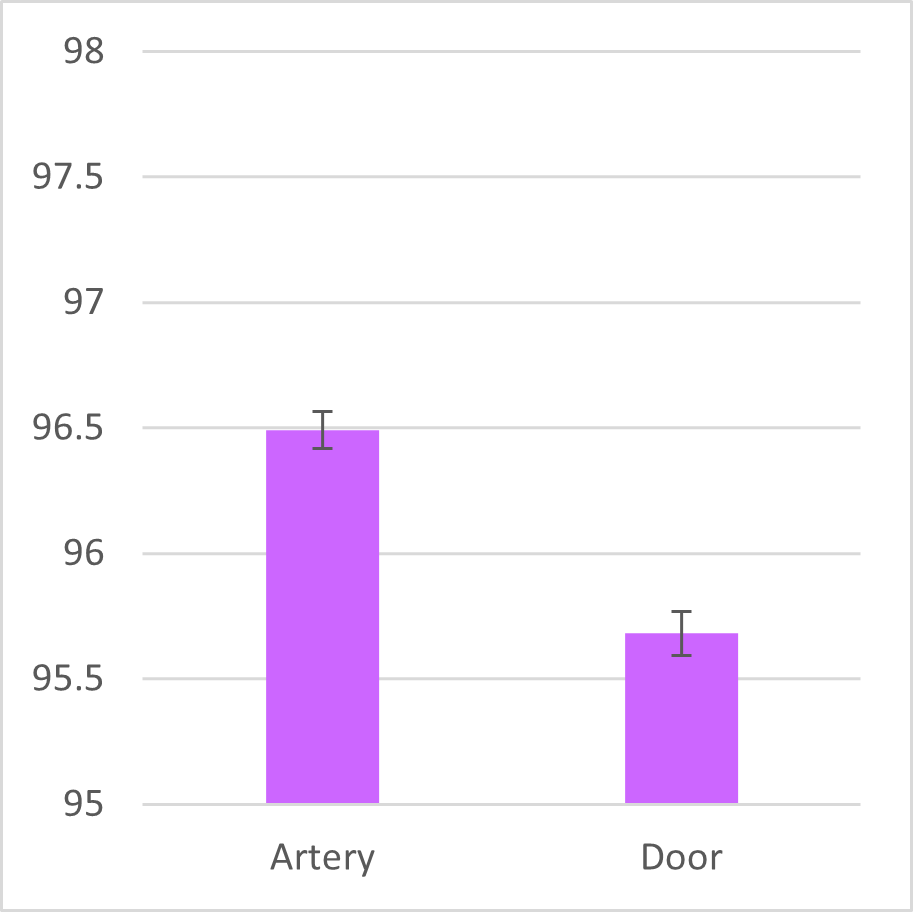}
\label{fig:2figsA}}
\qquad
\begin{minipage}{5cm}
\includegraphics[width=5cm]{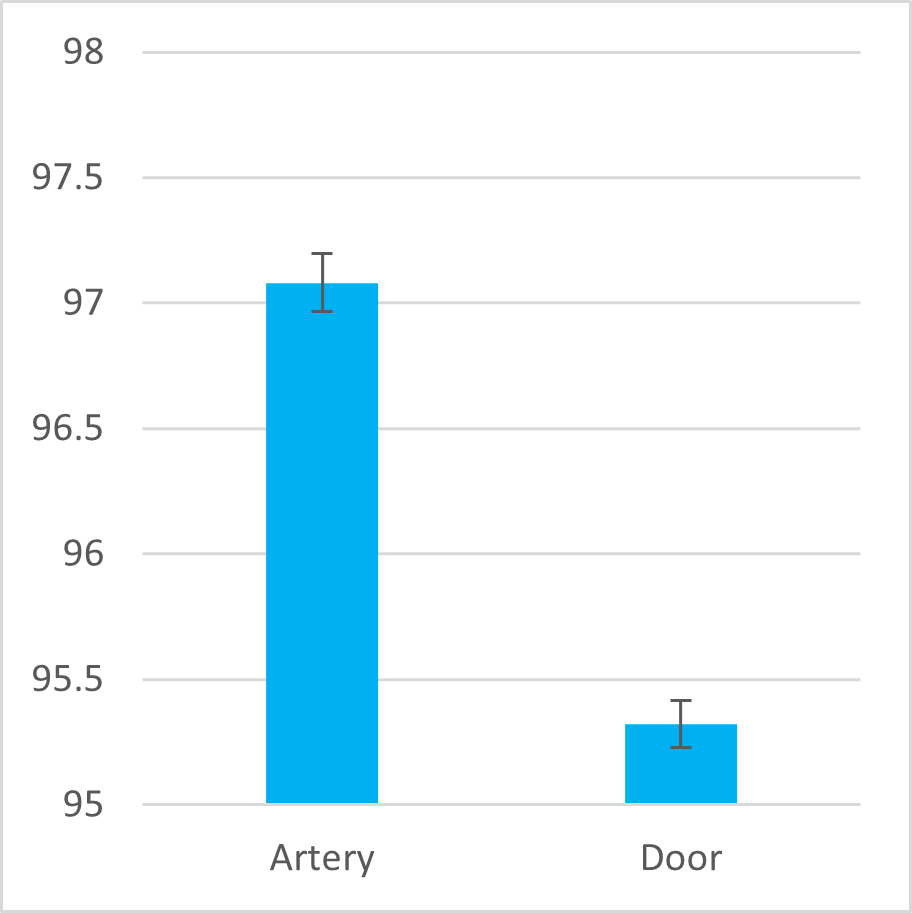}
\label{fig:2figsB}
\end{minipage}
\caption{Bar Plots of obtained median accuracy (\%) in the test dataset for the bi-encoder (left) and the BI+CE (right). The error bars display the standard deviation across the 5 runs.}
\end{figure}

\end{document}